\documentclass[10pt,journal]{IEEEtran}
\usepackage{graphicx,amsmath}
\usepackage[hidelinks]{hyperref}

\begin{document}
\title{Site-Conditioned Boundary Closure for Subsurface Seepage Forecasting and Relief Placement}
\author{Edward Holmberg, Elias Ioup, and Mahdi Abdelguerfi}
\maketitle

\begin{abstract}
Truncating a large heterogeneous diffusion model around a localized,
threshold-activated outlet saves degrees of freedom but severs exterior
feedback that governs when the outlet opens and how much it discharges.
By separating forcing-driven background head from local withdrawal
drawdown, Site-Conditioned Boundary Closure (SCBC) maps local physical
descriptors to positive exchange coefficients at the cut boundary.
Supervised on unforced donor-domain withdrawal responses, these
coefficients yield a passivity-preserving reduced operator that
generalizes across unseen forcing histories without refitting. We
evaluate SCBC on flood-driven subsurface seepage beneath levee
foundations against full-grid reference simulations (MODFLOW~6). Across
three held-out foundation panels, a single frozen SCBC tree estimator
reduced peak-discharge log RMSE by 26.1\%, 30.9\% and 42.6\% relative to
identical local patches using a shared boundary coefficient. We then
test whether better screening forecasts improve constrained
placement, minimizing peak landside toe head subject to a discharge
capacity limit and a budget of three sequential full-grid checks. On a
separate 36-foundation test using a coarser 30-state ring operator, a
neural SCBC estimator improved discharge forecasts across candidates yet
selected four over-capacity placements before checks, compared with one
for a shared-coefficient ring, which recovered all 144 reference optima
after three checks.
\end{abstract}


\section{Introduction}
\begin{figure*}[!t]
  \centering\includegraphics[width=.97\textwidth]{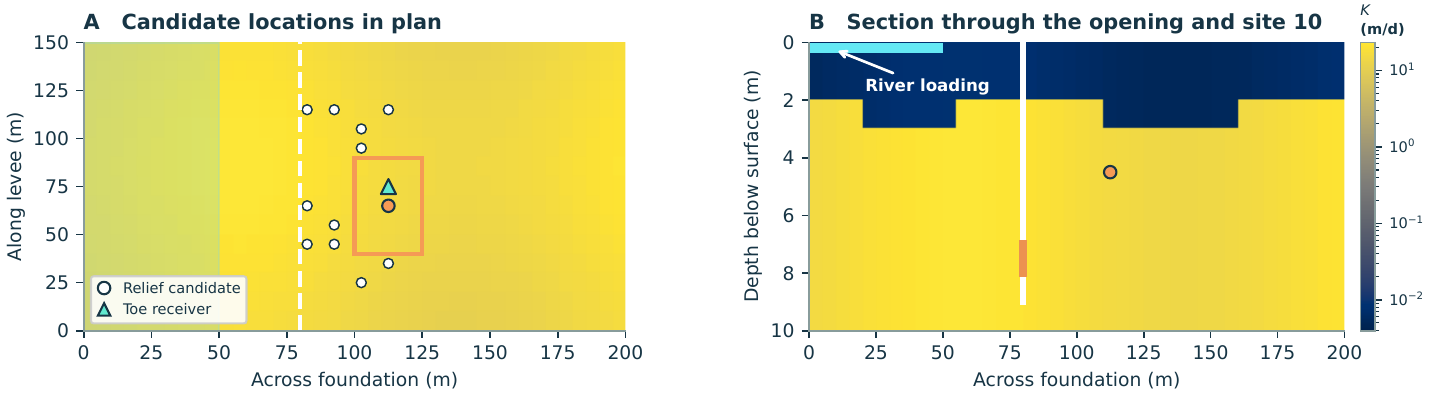}
  \caption{A held-out foundation used throughout the paper (3128). The maps
  show the prescribed synthetic conductivity field. (A) The horizontal slice is at
  4.5~m depth. Circles locate the twelve candidate relief sites in plan,
  including candidates screened at other depths. The orange outline bounds the
  patch for site 10. The triangle marks the toe-head receiver. (B) The
  section at 65~m along the levee passes through the relief outlet (orange point)
  and the opening (orange segment) in the buried cutoff (white). The receiver
  is one row away from this section. Cyan marks the river-loading footprint.
  Cross-section depth is exaggerated.}
  \label{fig:foundation}
\end{figure*}

During a flood, river water can move through permeable layers beneath
a levee and raise hydraulic head at the landside toe, the base of the
levee on its protected side. A passive seepage-relief outlet intercepts
some of this flow once local head exceeds the outlet elevation. Engineers must
weigh the reduction in peak toe head against the discharge the receiving
system must carry~\cite{mf6,usace}.
In Fig.~\ref{fig:foundation}, twelve candidate sites experience the
same flood and share the same opening in a buried cutoff. Their
conductivity, storage and hydraulic connections to the river vary.

Forecasting the discharge captured at one site requires accounting for
water drawn from beyond a small computational patch. In Site-Conditioned
Boundary Closure (SCBC), a regressor uses site properties to estimate
hydraulic exchange across the artificial patch boundary. The flow
equations then predict outlet discharge. We test whether estimating
exchange separately at each site improves forecasts on unseen foundations.
We also test whether that gain helps place a single relief outlet
when discharge capacity is limited and only a few sites can be checked.

\emph{Learning a boundary coefficient from hydraulic response.}
We separate the head driven by the river from the drawdown driven by
the outlet's own discharge~\cite{reilly,maddock}. Unit withdrawals on
donor foundations supply targets for learning effective exchange with
omitted cells. A new site's descriptors determine its coefficient before
its full-grid withdrawal response is checked. A common foundation-wide
calculation supplies river and receiver responses. At fixed properties and time step, the
local operator can be driven by different hydrographs without refitting
(Fig.~\ref{fig:patch}).

The answer to the first question was consistent across three panels.
At unseen relief sites, the same frozen $5\times5$ patch estimator reduced
peak-discharge log RMSE by 26.1\%, 30.9\% and 42.6\% relative to a
shared boundary coefficient (Table~\ref{tab:primary-patch}). An
exploratory $7\times7$ ridge fit reduced error by 55.1\% against its
own radius-matched control on the boundary-comparison panel.
All sites on a given foundation were held out together.

For a single relief placement, the result depended on which sites appeared
feasible. On a separate 36-foundation test, the neural ring improved
the discharge forecasts across candidates, yet selected four
over-capacity placements before hydraulic checks. The shared ring
selected one. After three sequential site checks the shared ring
recovered all 144 reference choices (Sec.~\ref{sec:decisions}). The
distinction matters in practice. A forecast error at a site with high
head relief can change the placement as soon as it moves that site
across the capacity limit.

\section{Subsurface Seepage, Boundary Truncation and Relief Placement}
\label{sec:setting}
\subsection{A diffusion operator with a local outlet}
Consider a network with nonnegative conductances assembled into a
time-invariant symmetric matrix $G$, positive diagonal storage $M$,
a prescribed load $f_t$, and a
single outlet at coordinate $e_s$. For a time step $\Delta t$, the
confined groundwater balance under backward Euler integration and
the constant-conductance MODFLOW~6 drain (\texttt{DRN}) boundary law are~\cite{mf6}
\begin{align}
 (D+G)h_t&=D h_{t-1}+f_t-e_s q_{s,t},\quad D=M/\Delta t ,
 \label{eq:balance}\\
 q_{s,t}&=C_d[e_s^\top h_t-z_d]_+,\quad [v]_+=\max(v,0).
 \label{eq:threshold}
\end{align}
Here $z_d$ is the outlet threshold (sill) elevation and $C_d$ is its
conductance. Positive $q_{s,t}$ denotes seepage discharge leaving the
foundation through the outlet at site $s$. Index $t=0$ denotes the end of
the first step after initial head $h_{-1}$. With a prescribed outflow
the state update is linear. The positive-part law closes the feedback
loop by setting the outflow from the resulting head.

\subsection{The exterior response a local model must approximate}
\label{sec:exterior-response}
Let $A=D^{-1/2}GD^{-1/2}$. For a withdrawal supported on patch cells
$p$, eliminating the exterior cells $e$ at shift $\sigma>0$ gives the
exact Schur complement~\cite{toselli,kron}
\begin{equation}
 S_p(\sigma)=A_{pp}+\sigma I-
 A_{pe}(A_{ee}+\sigma I)^{-1}A_{ep}.
 \label{eq:schur}
\end{equation}
If $w_s=(D^{-1/2}e_s)|_p$, then $w_s^\top S_p(\sigma)^{-1}w_s$
is the shifted head response at the withdrawal site. The exterior
term can couple distinct perimeter cells and varies with $\sigma$.
Our closures target this single input and readout through impulse
or shifted-response samples. Matching that scalar response leaves
the other entries of the exterior operator undetermined.

\subsection{Forecasting seepage and selecting a relief site}
Figure~\ref{fig:foundation} shows the physical realization. A flood
loads the river boundary, a buried cutoff wall impedes flow, and a
prescribed opening in the wall is common to all candidates. The passive relief outlet on the
protected side introduces the local head threshold in Eq.~\eqref{eq:threshold}. Let $u_{s,t}$ be its
head above sill with the drain omitted and $h^0_{\rm toe,t}$ the
corresponding toe head. A unit-rate withdrawal for one step gives
drawdown $k_{s,\ell}$ at the drain and $g_{s,\ell}$ at the toe at lag
$\ell$. By superposition~\cite{reilly,maddock}, prior withdrawals
lower the available head and the current outflow lowers it by
$k_{s,0}q_{s,t}$. Substituting the superposed drain head into
Eq.~\eqref{eq:threshold} gives
\begin{align}
 q_{s,t}&=\frac{C_d}{1+C_d k_{s,0}}
 \left[u_{s,t}-\sum_{\ell=1}^{t}k_{s,\ell}q_{s,t-\ell}\right]_+,
 \label{eq:outflow}\\
 h^{(s)}_{\rm toe,t}&=h^0_{\rm toe,t}
       -\sum_{\ell=0}^t g_{s,\ell}q_{s,t-\ell}.
 \label{eq:toe}
\end{align}
The initial outflow history is empty. Lag zero is the end of the pulse.
Our step is one day. Head is in metres, discharge in m$^3$/day and
$C_d$ in m$^2$/day. Both kernels give metres of drawdown per unit
discharge. A foundation-wide calculation supplies $u_{s,t}$,
$h^0_{\rm toe,t}$ and $g_{s,\ell}$. The site model estimates $k_{s,\ell}$.

For a flood, $Q_s=\max_t q_{s,t}$ is peak seepage discharge
captured at outlet $s$ and $H_s=\max_t h^{(s)}_{\rm toe,t}$ is peak
head at the landside toe receiver. The no-installation case has
$Q_\emptyset=0$. With receiving-system capacity $C$, the native
reference choice is
\begin{equation}
 s^*(C)\in\arg\min_{s\in\mathcal S\cup\{\emptyset\}:\,Q_s\le C} H_s .
 \label{eq:siting}
\end{equation}
The selection policy predicts head benefit $B_s=[H_\emptyset-H_s]_+$.
The two peak heads can occur on different days. A choice passes if native
discharge meets capacity and native peak head is within 0.001~m of the
best feasible head. Exact recovery also requires the reference action
index. For tied native heads we take the first feasible action in archive
order (no installation first, then ascending site index). Any capacity
exceedance fails the feasibility test.

\section{Related Work}
\subsection{Learning outputs, closure terms and interface conditions}
Groundwater surrogates have long been used for repeated simulation and
optimization~\cite{asher}. Taccari et al.~\cite{taccari} use a deep
operator network to predict hydraulic head for heterogeneous aquifers
and varying pumping locations. For levee seepage, Pokhrel et
al.~\cite{pokhrel4dvar} combined a surrogate trained on OpenFOAM
simulations of Richards' equation with adjoint-free variational data
assimilation to update a day-ahead forecast. In river hydraulics,
Holmberg et al. learned to forecast stage with a recurrent neural
operator~\cite{holmbergRecurrent} and later estimated how
water-surface increments persist at native HEC-RAS 2D
cells~\cite{holmbergInertia}. Here the learned quantity is the
exchange with cells omitted from the drain model. The local flow
equations use that exchange to compute discharge. Physical precedents include
the pumping response functions of Maddock and Lacher~\cite{maddock} and
Lanczos reduction of transient groundwater flow by Dunbar and
Woodbury~\cite{dunbar}.

Three related approaches clarify the role of the boundary. Mehl and
Hill~\cite{mehl} enforce consistency between coarse and refined
groundwater grids through interface head and flux. Gosses et
al.~\cite{gosses} explicitly represent boundary conditions in a
groundwater POD model, and Buhr et al.~\cite{buhr} describe the
construction and coupling of local reduction spaces. Taghibakhshi et
al.~\cite{interface} learn interface conditions for an iterative
domain-decomposition solver. Brugnolli et al.~\cite{hyperboundary}
encode boundary conditions in a learned operator. Ivagnes et
al.~\cite{ivagnes} learn corrections for unresolved fluid modes.
Equation~\eqref{eq:schur} eliminates the exterior at each shift.
The fitted coefficient approximates its effect on the retained
withdrawal response. We isolate the value of conditioning on the site by holding the
hydraulic solver fixed and comparing learned coefficients with a
single calibration shared across donors.

\subsection{Prediction for decisions under hard constraints}
Elmachtoub and Grigas~\cite{spo} distinguish predictive error from
the objective loss incurred when a forecast enters an optimization.
Mandi et al.~\cite{mandi} study decision-focused training when predicted
parameters affect feasibility. Our models are trained on hydraulic
responses. Capacity and head relief enter when the forecast is used
to select a drain, so their errors can be assessed independently of
the training objective. Gardner et al.~\cite{gardner} use the
probability of constraint satisfaction to guide sequential evaluation.
Our checked policies follow that principle using site-specific
hydraulic calculations.

A related question in spatial search is which candidates merit an
expensive evaluation. Ioup et al.~\cite{ioupMtree} used an M-tree
to filter candidates and reduce the number of road-network distances
evaluated in an aggregate nearest-neighbor query. The quantity used to rank sites also
matters. Holmberg et al.~\cite{holmbergGcoos} evaluated Gulf of Mexico
sensor locations by the coverage of a proposed observing network and
the robustness of that coverage. Here sites are ranked by predicted
peak head relief, and discharge capacity determines eligibility.

In groundwater engineering, Wang and Ahlfeld~\cite{wang} coupled flow
simulation to well placement and rate optimization for remediation.
Chen et al.~\cite{chenops} optimized active pumping at existing levee
relief wells. Here the action is to install one passive drain, or none.
Evaluating every candidate with MODFLOW~6 lets us measure both how the
prior classified the alternatives and what happened to the one selected.

\section{Site-Conditioned Hydraulic Exchange}
\label{sec:approach}
\subsection{A learned boundary for a truncated hydraulic patch}
\begin{figure*}[t]
 \centering\includegraphics[width=.985\textwidth]{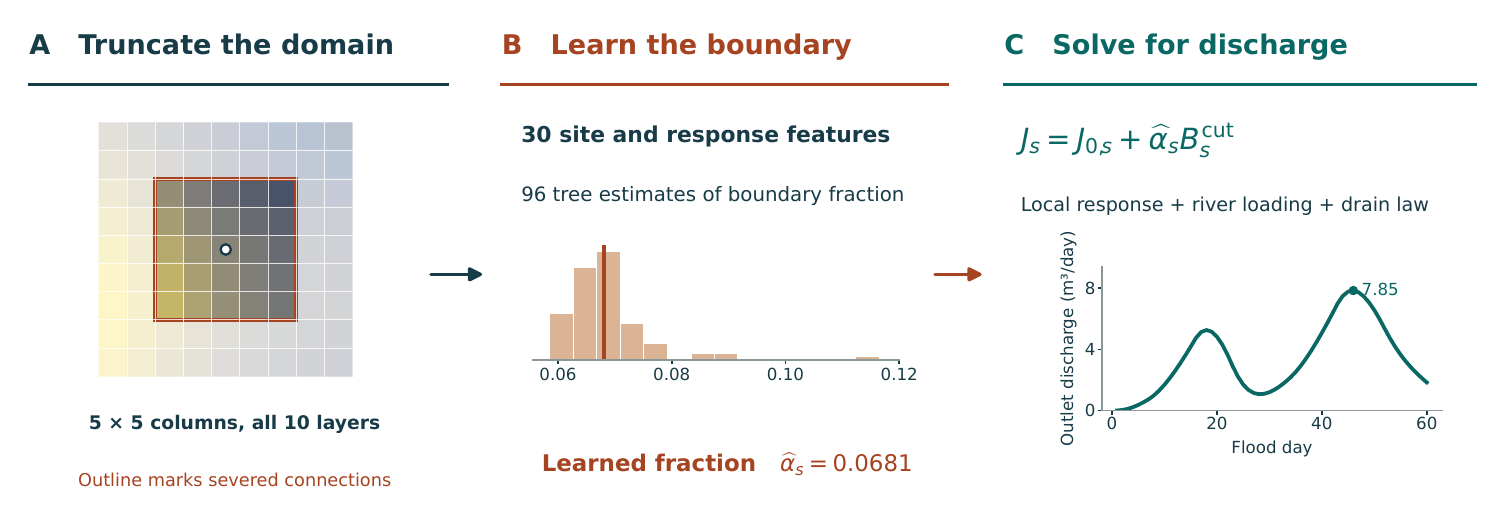}
 \caption{How a learned boundary coefficient produces a hydraulic forecast
 at site 10 of held-out foundation 3128. (A) The outline separates the
 retained patch from the faded exterior on the drain layer. The dot marks
 the drain. (B) The 96 trees estimate the fraction of severed conductance
 to retain. The vertical line gives their geometric mean,
 $\widehat\alpha_s=0.0681$. (C) The fraction scales the cut-face matrix
 in the hydraulic operator. Equations~\eqref{eq:kernel} and
 \eqref{eq:outflow} give the daily drain discharge shown, with a predicted
 peak of 7.85~m$^3$/day. Figure~\ref{fig:borderline} compares it with
 the native peak and the assigned capacity.}
 \label{fig:patch}
\end{figure*}

Figure~\ref{fig:patch} traces an SCBC forecast from the local foundation grid
to a candidate relief outlet's discharge hydrograph. The patch cuts a block of up to $5\times5$
columns, through all ten layers, from the foundation with its prescribed cutoff opening. Each cut
horizontal face loses a neighbor that would otherwise exchange water
with the patch. Retaining all severed conductances against a
zero-perturbation exterior would clamp the missing neighbors at their
reference heads. Dropping those conductances would insulate the cut.
The fitted fraction $\alpha_s$ specifies the strength of exchange
between these limits for the local withdrawal response.

Equation~\eqref{eq:schur} gives the exact exterior contribution.
We keep its effect on the site's withdrawal response through a fitted
fraction of the severed conductance.

Write $L_{0,s}$ for the storage-scaled patch operator with insulated cuts.
For patch index set $p$, let $c^{\rm cut} =
G_{pp}\mathbf{1}_{p}-(G\mathbf{1})|_{p}$. The principal submatrix
$G_{pp}$ retains the full diagonal conductances and drops the
off-diagonal entries for omitted neighbors. The physical boundary
conductance appears in both row sums and cancels, leaving the total
conductance from each retained cell to its omitted neighbors. We scale
these cut conductances by storage to form
$P_s=\operatorname{diag}(c^{\rm cut}_i/D_{ii})_{i\in p}$.
For a local basis $U_s$,
\begin{align}
 J_{0,s}&=U_s^\top L_{0,s}U_s,\quad B_s^{\rm cut}=U_s^\top P_sU_s,\nonumber\\
 J_s(\alpha)&=J_{0,s}+\alpha B_s^{\rm cut}.\label{eq:patch}
\end{align}

Polynomial Krylov projection has a groundwater precedent~\cite{dunbar},
and rational Krylov methods enrich a projection space with shifted
solves~\cite{gugercin}. We start from the patch coordinate vector $e_s|_p$ at the drain and
use it as the right-hand side of each shifted solve. With diagonal
storage, this vector is proportional to the restricted withdrawal
vector $w_s$ in Sec.~\ref{sec:exterior-response}.
Our basis uses four polynomial directions from
the clamped operator and three local shifts (1, 0.10 and 0.012) at each
of two fractions (0.012 and 1). After normalizing these ten candidate
vectors, column-pivoted QR~\cite{businger} retains columns with
$|R_{ii}|>10^{-11}$.
The resulting basis has at most ten directions and is shared by the learned
and constant versions. These six solves act on the local patch. Training and
site checking use a separate full-grid response.

The reduced withdrawal vector is $v_s=U_s^\top w_s$.
Projecting Eq.~\eqref{eq:balance} onto $U_s$ gives the one-day propagator
$(I+J_s)^{-1}$. A unit-rate withdrawal lasting one time step applies
the propagator once at the end of that step and once per subsequent
lag. The local pulse response, for lag $\ell=0,\ldots,59$, is
\begin{equation}
 k_{s,\ell}(\alpha)=v_s^\top(I+J_s(\alpha))^{-(\ell+1)}v_s .
 \label{eq:kernel}
\end{equation}
For $J_s=V\Lambda V^\top$, each eigenvalue contributes
$(1+\lambda_j)^{-(\ell+1)}(V_{:j}^\top v_s)^2$. This evaluates all
lags from one reduced eigendecomposition. Summing pulse coefficients
over successive lags gives the sustained-withdrawal response used in training.

\subsection{What the regression learns}
For each donor site, the full-grid six-state projection supplies a target
withdrawal response $k^*$. Define $S_s(h;\alpha)=\sum_{\ell=0}^{h-1}
k_{s,\ell}(\alpha)$ and define $S_s^*(h)$ analogously. We choose the
site label on a logarithmic grid $\mathcal A$ of 25 fractions spanning
$[0.012,1]$ by
\begin{equation}
 \alpha_s^*=\arg\min_{\alpha\in\mathcal A}
 \frac{1}{5}\sum_{h\in\{1,4,12,32,60\}}
 \left(\log\frac{S_s(h;\alpha)}{S_s^*(h)}\right)^2 .
 \label{eq:label}
\end{equation}
The shared control minimizes this same donor-averaged loss over
the 25 fractions in $\mathcal A$.
For the site-specific rule, the input vector contains 13 site descriptors,
seven patch summaries and ten measurements of how the two limiting
patch models respond. The 13 common inputs are layer, row, column, lateral
distance to the toe row, lateral and vertical distances to the opening,
column offset from the cutoff, and logs of six hydraulic quantities:
cell conductivity, storage per step, diagonal conductance, drain
conductance, mean neighboring conductivity and mean reciprocal
neighboring conductivity. Neighborhood means use a clipped
$3\times3\times3$ window. Distances use grid indices.

The seven patch inputs are logs of arithmetic, harmonic and
root-mean-square conductivity, total cut conductance, total storage
per step, and mean conductivity and storage per step over cut nodes.
The ten response inputs are $\log S_s(h;1)$ and
$\log[S_s(h;0.012)/S_s(h;1)]$ at each of the five horizons.
A regressor estimates $\log\alpha_s^*$. Exponentiation with clipping
keeps the estimate within the grid limits while allowing intermediate values.

Training uses 324 sites from 27 donor foundations. For each held-out
donor foundation, ridge regression~\cite{hoerl} and extremely randomized
trees~\cite{geurts} fit on the other 26. Grouping sites by foundation
preserves the unit to which we seek transfer, following the principle of
blocked validation for structured data~\cite{roberts}. Candidate models are scored
by taking the square root of the mean held-out hydraulic loss in Eq.~\eqref{eq:label}, using the nearest fraction on the log grid to reuse the precomputed
five-horizon responses. The selected ensemble has 96 trees, minimum leaf size five,
85\% of features considered per split and random seed 731. It is refit
on all donors. Deployment evaluates its continuous predicted fraction.
Section~\ref{sec:donor-sensitivity} checks sensitivity to this
scoring convention and to the tree seed.

\subsection{A compact neural model on hydraulic rings}
\begin{figure*}[t]
 \centering\includegraphics[width=.985\textwidth]{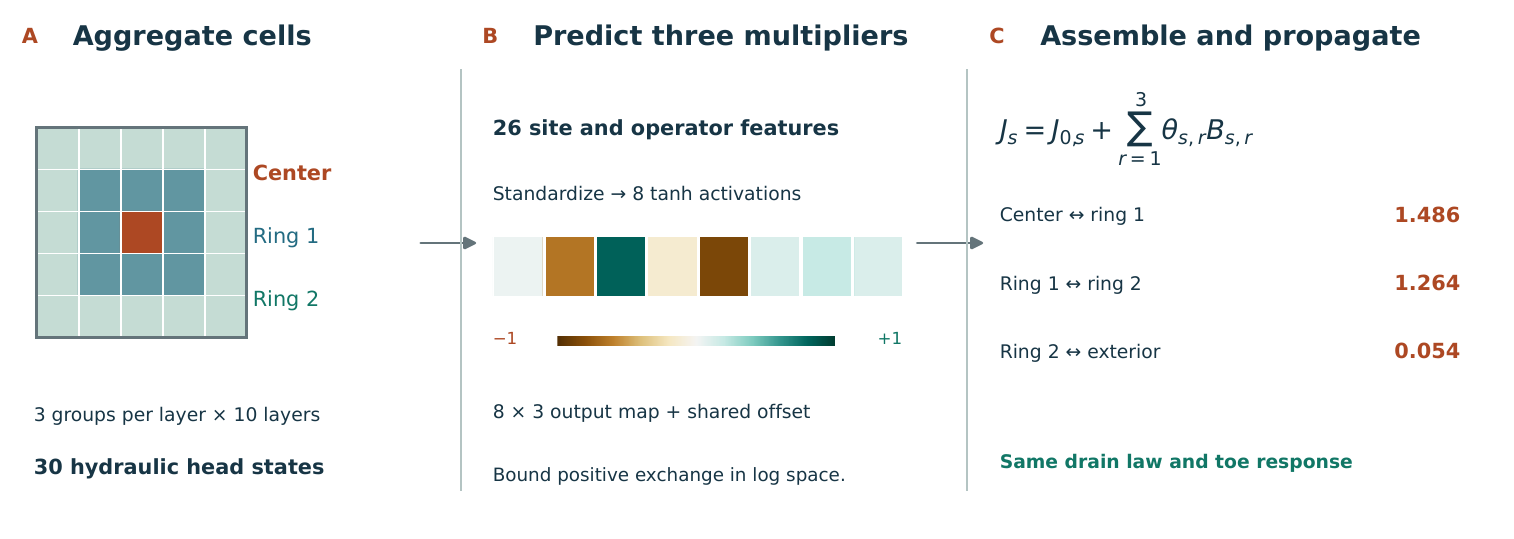}
 \caption{Neural ring architecture. (A) Group membership depends on
 horizontal distance from the drain at each layer. (B) The network maps
 26 descriptors into three offsets around the shared calibration. The eight tiles show the signed hidden activations for this site, in unit order. (C)
 Each multiplier acts on a specified exchange matrix. Values are the
 actual predictions for the same drain as in Fig.~\ref{fig:patch}.
 Hydraulic time integration and toe readout are common to both architectures.}
 \label{fig:ring}
\end{figure*}

Figure~\ref{fig:ring} shows a coarser hydraulic aggregation of the same local neighborhood.
At each depth, the center, the eight cells at a horizontal grid distance
of one in the maximum norm, and the sixteen at distance two form three
aggregates on an interior patch. The sums of storage and face
conductance define a 30-state operator through the ten layers.
Truncated edge patches use the available cells. Vertical exchange and
physical boundary terms enter $J_{0,s}$. Three matrices represent
center--inner, inner--outer and outer--exterior exchange:
\begin{equation}
 J_s(\boldsymbol\theta)=J_{0,s}+\sum_{r=1}^{3}\theta_{s,r} B_{s,r}.
 \label{eq:ring}
\end{equation}
With the projected source replaced by the storage-scaled center
aggregate at the drain depth, Eq.~\eqref{eq:kernel} also evaluates this
model's response.

The ring uses the same 13 common features. Five additional features
summarize conductivity: arithmetic and harmonic means for the center
column and for the patch, and the patch standard deviation plus 0.001.
Eight further quantities describe the assembled operator: four relevant
diagonal entries at the drain depth (retained, center--inner,
inner--outer and exterior terms), total cut conductance, and the three
aggregate storage masses at that depth. The five and eight feature groups
enter on the log scale with the positivity safeguards in the implementation.
Standardization uses donor means, a minimum scale of 0.10, and clipping
to $[-6,6]$.

For standardized descriptors $x_s$, the three network outputs are
$a_s=b_0+W_2^\top\tanh(W_1^\top x_s+b_1)$. The shared control learns
the three intercepts $b_0$. The network holds these fixed and learns
offsets. For component $r$, set
\begin{equation}
 \log\theta_{s,r}=\log l_r+
 \frac{\log u_r-\log l_r}{1+\exp(-a_{s,r})},
 \quad r=1,2,3 .
 \label{eq:ring-logistic}
\end{equation}
Here $l=(0.05,0.03,0.008)$ and $u=(2,2,1.5)$. All three positive
multipliers may exceed one. Each rescales a conductance in the
aggregated operator, including the outer boundary term.
The network has eight
hidden units and 240 adjustable parameters beyond $b_0$.

Both ring variants fit the response
$\rho_s(\sigma)=v_s^\top(\sigma I+J_s)^{-1}v_s$ at
$\sigma\in\{0.015,0.06,0.25,1,4\}$. The loss is the mean squared
log ratio of the predicted response to its six-state teacher target,
with $6\times10^{-5}$ times the squared norm of the network parameters
added for the learned offsets. The 216 five-shift target vectors come from twelve sites
on each of 18 donor foundations. Nine other donors calibrate ranking
uncertainty. The L-BFGS-B optimizer~\cite{byrd} reached the prescribed 85-iteration limit for
the neural fit. Reported results use that fitted iterate. The site
fraction and three multipliers are effective parameters of these
discretizations. We evaluate them by the withdrawal responses and forecasts they produce.

\subsubsection*{What the closure preserves}
In storage-scaled coordinates, each internal edge contributes a
nonnegative weight times an outer product of a cell-difference vector;
a boundary conductance contributes a nonnegative diagonal term.
Projection and aggregation preserve these positive semidefinite forms.
Both $J_s(\alpha)$ and $J_s(\boldsymbol\theta)$ therefore remain
symmetric positive semidefinite at every admissible coefficient value.
For the local withdrawal subsystem write
$(I+J)x_t=x_{t-1}+v a_t$, with withdrawal input $a_t$ and collocated
drawdown $y_t=v^\top x_t$. Multiplication by $x_t^\top$ gives
\begin{equation}
 E_t-E_{t-1}
 =a_t y_t-x_t^\top Jx_t-\tfrac12\|x_t-x_{t-1}\|^2,
 \quad E_t=\tfrac12\|x_t\|^2 .
 \label{eq:local-energy}
\end{equation}
Thus this subsystem is passive for the specified input and readout,
and the unforced step cannot increase $E_t$. These properties hold for
any fitted coefficients within the stated bounds. Peak-discharge and
placement accuracy are evaluated separately against MODFLOW~6.

\subsection{From twelve forecasts to one checked placement}
\label{sec:decision-loop}
\begin{figure*}[!t]
  \centering\includegraphics[width=\textwidth]{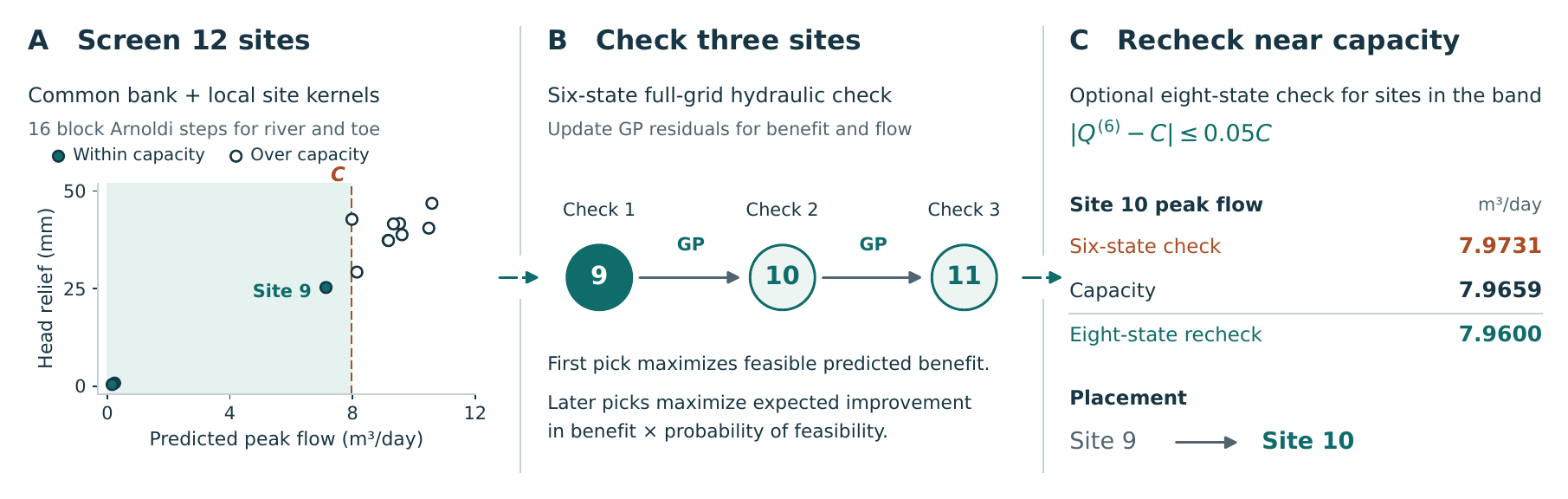}
  \caption{From twelve candidate forecasts to one placement. (A) The common
  river and toe response bank and the local withdrawal kernels supply the
  neural-ring forecasts on foundation 3128. Filled circles are predicted to meet capacity, and open circles to exceed it.
  (B) Site numbers show the recorded query order. The first pick uses the
  screening forecasts. After each six-state check, the Gaussian process
  (GP) uses the residuals to choose the next site while budget remains.
  (C) In the retrospective refinement, two of the three checked sites fell
  within 5\% of capacity $C$ on the six-state estimate. Rechecking them
  changed the choice from site 9 to site 10. The same rule was later frozen
  for the boundary challenge.}
  \label{fig:decision-loop}
\end{figure*}

Figure~\ref{fig:decision-loop} traces a three-check decision on a
confirmation foundation, from the twelve initial forecasts to the
placement selected after a near-capacity recheck. For every candidate, the workflow computes the river-stage response at
the site and the toe response to a site withdrawal. The accelerated
implementation builds a common basis from a river input and a reciprocal
toe input using 16 block Arnoldi steps on the implicit time propagator. Each new
two-column block is orthogonalized twice against the existing basis
and factored by QR. Only candidate and toe rows are needed for readout. In this confined discretization the input--output kernel
$D^{-1/2}(I+A)^{-(\ell+1)}D^{-1/2}$ is symmetric, so the response at a
candidate to a pulse at the toe equals the toe response to a pulse at
that candidate. This is the reciprocity used in the computation~\cite{falade}. This setup is shared across candidates. The site kernels
and common responses yield the predicted peaks in Eqs.~\eqref{eq:outflow}
and~\eqref{eq:toe}.

\begin{table*}[!t]
\centering\caption{Progression of the five held-out tests. Each site appears
under one flood in its panel. Near-capacity choices on the same foundation
reuse the same native simulations. Each panel used fresh foundations.
The confirmation recheck, boundary-comparison radius refits and three-check
replay were added after their respective panel scores. The boundary
challenge tested the frozen recheck prospectively.}
\label{tab:design}
\small
\begin{tabular}{lrrrrp{.30\textwidth}}\hline
Panel & Foundations & Candidates & Near-limit decisions & Native runs & Scientific purpose\\ \hline
Initial & 36 & 432 & 144 & 468 & Test the learned patch against a shared fit and an accelerated five-check selector.\\
Confirmation & 27 & 324 & 108 & 351 & Repeat the patch comparison and test neural aggregation, including nine material-stress realizations.\\
Boundary challenge & 9 & 108 & 36 & 117 & Stress feasibility decisions within 1\% of the designated hydraulic flow estimate.\\
Boundary comparison & 18 & 216 & 72 & 234 & Replicate the patch forecast and compare the training target, estimator and exterior treatment.\\
Ring estimator test & 36 & 432 & 144 & 468 & Hold the hydraulic ring fixed and compare four estimators of its three conductance multipliers.\\ \hline
\end{tabular}
\end{table*}

A hydraulic check recomputes the local response using the whole foundation
operator. The six-state checker starts with four polynomial Krylov
directions, adds a shift-one inverse direction, and refines a shift-0.1
direction with up to two preconditioned conjugate-gradient iterations.
The shift-one solution initializes the iteration, and $(A+I)^{-1}$
preconditions its residuals using the existing time-step factorization. It serves as both the
development teacher and the deployment check. Independent native
simulations score the final action. The first check takes the
predicted-feasible site with the greatest predicted benefit, or the
smallest predicted discharge if all sites appear infeasible. We condition
two independent Gaussian-process (GP) residual models on the checked
minus prior log predictions for head benefit and peak flow~\cite{gpml}, adding $10^{-7}$~m to benefit and $10^{-6}$~m$^3$/day to flow before taking logarithms. Their covariance combines
Mat\'ern-$3/2$ kernels on pairwise distances between candidates in three feature groups:
coordinates, site features (including material and forcing summaries),
and six hydraulic-link features. The link features are constructed for
each candidate with the first checked site as the common anchor. Each subsequent query
maximizes expected head-benefit improvement multiplied by the estimated
probability of satisfying capacity. This is a constrained acquisition
criterion related to that of Gardner et al.~\cite{gardner}. The
recommended drain has the largest positive checked benefit among checked
sites judged feasible. If none has positive feasible benefit, the
policy recommends no installation. The optional refinement reevaluates
each checked site whose six-state peak is within $0.05C$ of capacity
using eight states, then repeats the selection over the same shortlist.

\section{Experimental Design and Numerical References}
\label{sec:design}
\subsection{Foundations, forcing and native references}
The meshes are 200~m across the levee, 150~m along it and 10~m deep.
Forty columns, fifteen rows and ten layers give 6,000 confined cells,
each $5\times10\times1$~m. The nine design strata cross continuous,
pinching and channelized material structure with nominal conductivity
contrasts of 30, 300 and 1,500. A spatially varying fine blanket overlies
more permeable material. Some realizations also contain a deeper lens.
The cutoff at 80~m penetrates five, seven or nine layers. A prescribed
opening restores one internal face. A general-head river boundary acts
on the upper layer of the first ten columns. Landward boundary and
initial heads are zero relative to the model datum. The receiver is
at 112.5~m across the foundation and 4.5~m depth.

The twelve candidate screens occupy four columns between the cutoff
and toe, with three lateral bands per column. Screen depths and the
opening position vary across realizations. Each candidate seepage-relief outlet is represented by one optional
constant-conductance MODFLOW~6 \texttt{DRN} cell with sill elevation
zero relative to the model datum. Within a foundation
the same $C_d$ is assigned to each candidate. The generator sets it to
the conductance gained by restoring the specified cutoff face. This
choice couples drain strength to the foundation design. An external test
could vary outlet conductance independently.

All five panels use the U.S. Geological Survey release of MODFLOW~6.8.1~\cite{mf6release} with its Groundwater Flow Model~\cite{mf6}. The native model has
confined storage (water released as head declines in saturated cells),
node-property flow and a horizontal flow barrier. A general-head boundary exchanges water in
proportion to the difference between prescribed boundary head and cell head. Each synthetic hydrograph is specified by
piecewise-linear stage knots, sampled daily, one backward-Euler step per
day. The archived events span 48--60 daily steps. Baseline and candidate trajectories on a foundation share all stage,
opening, material and initial conditions. We use the native head observation
at the toe and the outward-positive volumetric discharge from the
candidate's \texttt{DRN} budget to form the two event peaks.
Thirteen simulations per foundation cover no installation and each of the
twelve candidates. Once computed, those thirteen outcomes score every
capacity without generating additional reference simulations. Across the first four panels, all 1,170 saved trajectories completed under the first solver configuration according to the archived retry flags.

\subsection{What was fitted, frozen and held out}
Table~\ref{tab:design} separates the successive design decisions.
The 27 fitting donors comprised three realizations in each of the nine
design strata. The two donor hydrographs used for peak targets and GP
calibration were synthetic stage sequences of 44 and 54 daily steps.
The original patch labels used withdrawal pulses alone.
The first panel used four new flood shapes and four realizations per
design stratum. The second used another four shapes across 18 standard
and nine material-stress foundations. In the stress group, cells with
conductivity below 0.8~m/day had conductivity multiplied by 0.18 and
specific storage by 1.8. Other cells received factors of 1.6 and 0.65.
The boundary challenge used nine new high-contrast realizations and three new
flood shapes. Each realization received one prescribed event. The panels
were designed in succession. The neural ring and the selective recheck
entered evaluation after the results that motivated them.

Three ordinary capacities were fixed at 2, 10 and 25~m$^3$/day.
For the first two panels, a previously fitted ridge predictor located
its highest-benefit site ignoring capacity. Four near limits were 0.80,
0.95, 1.05 and 1.20 times its predicted peak flow at that site. On the
boundary-challenge panel, the same selection procedure nominated a site whose full-bank
six-state check set the scale. Capacity was then 0.99, 0.9975, 1.0025
or 1.01 times that value. The accompanying archive includes the protocols and hashes of the frozen selection inputs. These limits were assigned before opening the native outcomes. The boundary challenge deliberately concentrates cases near a
decision boundary. Its pass rate describes that test design.

\subsection{Controls and equal-budget comparisons}
Table~\ref{tab:models} distinguishes the model that supplies the initial
forecasts from the GP that selects which forecasts to check.

\begin{table*}[!t]
\centering\small
\caption{Models available before the first check. Donors are training
foundations. Shared coefficients are fitted once and used at all sites.
A check replaces the local withdrawal response with a full-grid six-state
calculation. The selective eight-state recheck applies only to queried
finalists near capacity.}
\label{tab:models}
\renewcommand{\arraystretch}{1.13}\setlength{\tabcolsep}{4.4pt}
\begin{tabular}{@{}llll@{}}
\hline
Hydraulic model & Quantity supplied & Donor target & Estimator (donor foundations) \\ \hline
$5\times5$ Krylov patch & One boundary fraction & Cumulative pulse at five horizons & Shared, trees, ridge (27) \\
$5\times5$ peak-target variant & One boundary fraction & Peak flow on two donor floods & Shared, trees, ridge (27) \\
$7\times7$ truncated patch & One boundary fraction & Cumulative pulse at five horizons & Shared or ridge (27) \\
$7\times7$ with coarse exterior & Assembled exterior terms & Fine-grid cell properties & Deterministic assembly \\
30-state ring & Three exchange multipliers & Local response at five shifts & Shared, affine, trees, neural (18) \\
Direct ML control & Two log corrections to ring peaks & Checked benefit and flow & 100-tree regressor (18) \\
Conventional GP prior & Log corrections to one-cell forecasts & Checked benefit and flow & Ridge (27) \\
\hline\end{tabular}
\end{table*}

The patch ablation holds the local domain, basis, descriptor extraction
and hydraulic rollout fixed for estimation of discharge. The ring
ablation holds aggregation, training foundations and target shifts
fixed. GP uncertainty enters only after screening, when the policy selects sites to check. The direct ML control uses the same 18 training
foundations as the ring network. It has access to the ring descriptors,
forcing-dependent features and shared-ring peak forecasts. It predicts
log corrections to peak head benefit and peak flow, supervised by checked
outcomes under the two donor flood hydrographs. Ridge and tree alternatives were
compared using three foundation-grouped folds. This additional
control selected a 100-tree regressor. Its inputs and peak supervision
differ from the ring network's, so the comparison measures the two
complete predictors. The shared-versus-neural ring ablation isolates the site-specific offsets
at fixed hydraulic representation for the pre-check forecasts.

The conventional GP starts with a one-cell approximation to the drain's
withdrawal response. Let $r_s=(1+G_{ss}/D_{ss})^{-1}$. Keeping only
the drain cell in the withdrawal calculation gives the pulse response
$k^{(1)}_{s,\ell}=r_s^{\ell+1}/D_{ss}$. We use it in
Eqs.~\eqref{eq:outflow} and~\eqref{eq:toe} with the common river and
toe responses to obtain peak head benefit and peak discharge. Ridge
regression corrects each peak on the log scale using the difference
between the checked and one-cell log peaks on donor foundations.
Before taking logs, we add $10^{-7}$~m to each benefit and
$10^{-6}$~m$^3$/day to each discharge.

The conventional GP and each $5\times5$ patch GP use residuals on
all 27 donors under both donor floods. For each held-out donor the
conventional ridge prior is fitted on the other 26. The shared and
tree patch forecasts likewise use fractions estimated with that donor
excluded. On these out-of-fold residuals the covariance length scale,
mixture weights, noise and output variance are selected for each output.
The patch models inherit donor feature scaling from the conventional GP.
The deployed patch estimators then use all 27 donors. For the original
ring comparisons, 18 donors fit the predictors. The remaining nine set
prior-specific residual variances using the conventional GP's kernel
shape. The wider-patch replay reuses the frozen learned $5\times5$ patch GP,
and the matched ring-estimator replay gives all four arms the archived
neural-ring GP. Evaluation foundations contributed no calibration residuals.

The full-bank, two-channel and accelerated conventional GPs share the
ridge prior, acquisition rule, and six-state checker. The full bank
propagates three inputs (river, reciprocal toe head and reciprocal
section flow) over the full grid. The exact two-channel bank propagates
the river and toe inputs by the same full-grid step solve and stores
only requested receiver rows. The accelerated version approximates these
two channels by the common block projection. The matched zero-check analysis compares
the initial forecasts. At positive budgets all checked policies choose
from checked sites and no installation. Following a post-hoc feasibility
diagnosis on confirmation, a frozen variant enriched each checked site
within 5\% of capacity with two further orthogonalized inverse-step
directions. The list of checked sites was held fixed. The challenge
scores this eight-state rule prospectively against new native trajectories.

\begin{figure*}[t]
 \centering\includegraphics[width=.94\textwidth]{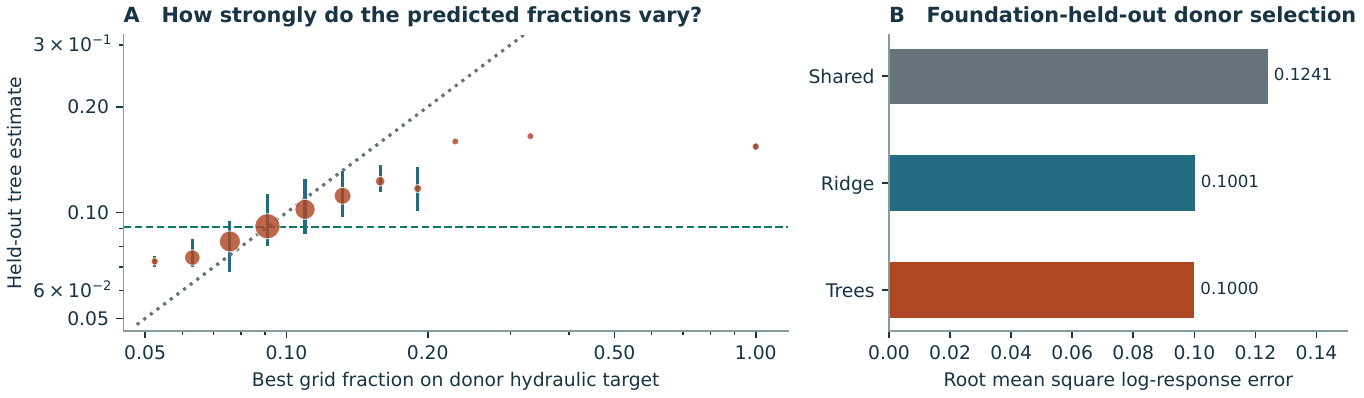}
 \caption{Training diagnostic on 324 donor sites. Each site is predicted
 with its foundation excluded from fitting. (A) For each grid-optimal
 fraction, the point is the median predicted fraction, the segment spans
 the 10th--90th percentiles and area increases with count. The dashed
 line is the shared fraction and the dotted line is equality. (B)
 Five-horizon hydraulic losses used for model selection. The ridge and tree losses differ by less than 0.0001. The independent
 foundation panels provide the deployed flow scores.}
 \label{fig:donor}
\end{figure*}

\subsection{Endpoints, dependence and computational accounting}
For $N$ candidate sites, initial peak-discharge error is
\begin{equation}
 R_Q=\left[\frac1N\sum_s\left(\log
 \frac{\widehat Q_s+\epsilon}{Q_s+\epsilon}\right)^2\right]^{1/2},
 \quad \epsilon=10^{-6}~\mathrm{m^3/day}.
 \label{eq:error}
\end{equation}
We also report mean absolute peak-discharge and head-benefit errors in
physical units. For paired differences in $R_Q$, we resample whole
foundations with replacement 10,000 times, preserving all twelve sites
and matching model and control in each draw. We report percentile
intervals conditional on the saved fits. This is a paired cluster
bootstrap with the foundation as the resampling unit~\cite{fieldwelsh}.
Intervals described as descriptive condition on fixed fits and carry no
multiplicity adjustment. For the two primary ring-estimator contrasts
we use Bonferroni-adjusted intervals as specified below.
Foundation win
counts and the zero-check placement rule were calculated retrospectively.
That rule chooses the greatest positive predicted benefit among sites
within predicted capacity and otherwise chooses no installation.

\begin{figure*}[!t]
\centering
\includegraphics[width=.94\textwidth]{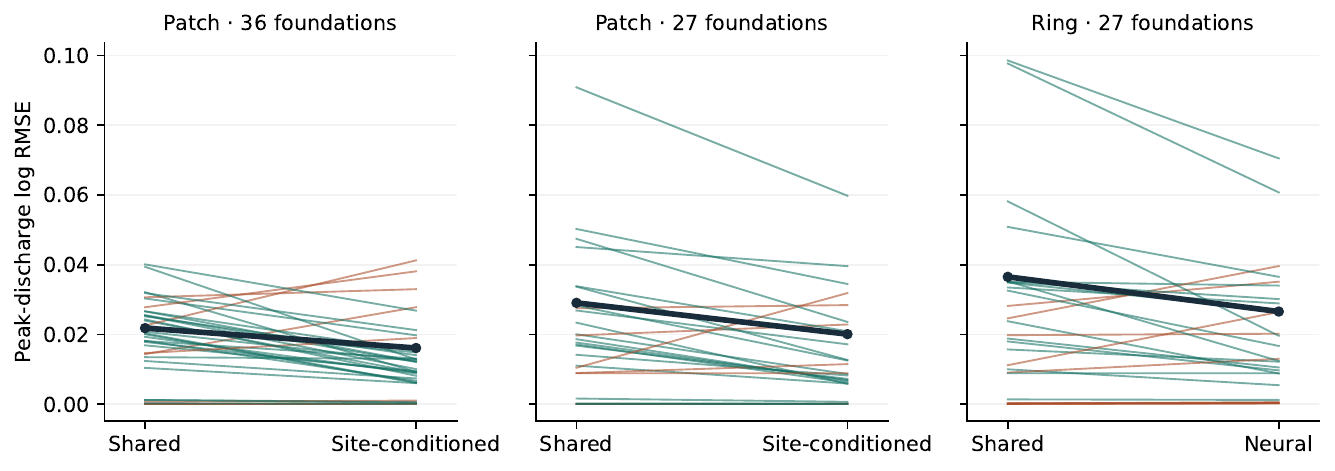}
\caption{Each thin line connects the foundation-level peak-discharge log RMSE, computed across its twelve candidates, for the two models. A descending line means the site-conditioned closure improved. The heavy line gives pooled log RMSE. The center and right panels use the same 27 foundations.}
\label{fig:paired}
\end{figure*}

Cold timing includes full-grid assembly, common response preparation, all
candidate features and predictions, queries and selection. Repeated
timings randomize method order within paired workloads. Ratios are
computed on matched runs before summarizing by foundation. The
initial series timed the first nine initial-panel foundations with three repetitions. Both confirmation series timed the same nine selected foundations with four repetitions each. The initial-panel run record specifies Python~3.12.14 and single-threaded OpenBLAS. Reported clock times describe this implementation and grid size.
Native runtime is outside the timed policy and appears only in the
construction of the reference data.

\subsection{Wider-patch and learning controls}
The boundary-comparison panel added nine standard and nine material-stress
foundations. Two new flood histories (53 and 58 daily steps) alternated
across the 18 foundations. Before scoring the 234 new native runs,
the archived study protocol specified the original $5\times5$ shared, frozen
tree and ridge estimators and the deterministic exterior. Its primary ML
contrast was between trees trained on peak discharge and trees trained
on the accumulated withdrawal response. All fits used the original 27 donor
foundations. For the alternative target we selected the fraction among the
same 25 candidates that best reproduced the six-state teacher's peak drain
discharge on two donor floods. Shared, ridge and tree estimators were fitted
with the same inputs available to their respective kernel-target controls.
The deterministic alternative retained fine-grid cells throughout a
$7\times7$ patch and grouped the surrounding cells in $4\times4$ horizontal
tiles within each layer, summing storage within each aggregate and the fine-face conductances
across each pair of aggregates. The patch and exterior shared the original river and toe
response calculations with all competing methods.

After examining these scores, we recalibrated the truncated $7\times7$
patch on the donors. For the shared fraction we minimized the mean
squared log discrepancy in accumulated response at five horizons. For
the site-specific rule, standardized ridge regression with the original $5\times5$ penalty of 24
predicted each site's log fraction from 30 descriptors, using the same
25-value grid to construct donor labels. Both rules used a direct local
transient solve. We also replayed the original three-check policy with
the patch forecasts and a common posterior calibrated for the original
learned patch. For the four near-capacity limits on each new foundation,
the original ridge prior and accelerated response bank set the threshold
scale before native scoring. First-screen timings covered full-grid setup
and all twelve forecasts on the nine new standard foundations (three
randomized repeats), excluding hydraulic checks.

\subsection{Matched estimators for the three ring coefficients}
A fifth panel tested whether the choice of estimator matters when every
candidate uses the same 30 hydraulic states. We held out 36 new
foundations. Each of the nine geological strata contributed two standard
and two stress-transformed realizations. Four new synthetic flood histories
were each assigned to nine foundations. The 432 candidate placements and
36 no-installation baselines supplied 468 native trajectories.

We kept the ring aggregation, 26 descriptors, normalization, coefficient
bounds and five-shift hydraulic targets fixed. Against the original shared
intercepts and archived eight-unit network we tested two estimators of
site-specific logit offsets. The first was an affine function with the same
quadratic weight penalty as the neural head. The second used up to 85
depth-two boosted regression trees, with at least eight donor sites per
leaf and a learning rate of 0.15. Each tree split used the negative
gradient of the hydraulic loss and each leaf used a damped Gauss--Newton
update. All four heads used the same 216 donor sites. Both the affine
and neural fits had an 85-iteration optimization budget. Two additional
prespecified neural initializations assessed forecast sensitivity. The
original fit remained the primary network.

We recorded all twelve site forecasts and the decisions under four
near-capacity limits before opening native scores. Every arm used the
original neural-ring GP uncertainty parameters, hydraulic checker and
three-check budget. The primary outcome was peak-discharge log RMSE.
We required a lower pooled error than each simpler site-dependent head
and an upper interval endpoint below zero for both paired differences.
For the two primary comparisons we used 97.5\% two-sided percentile
intervals, resampling whole foundations 10,000 times.

All 468 native reference trajectories completed. One candidate on foundation
87501 required the numerical retry procedure.

\newpage
\section{Results}
\subsection{Site-conditioned exchange improves discharge forecasts on three panels}
\label{sec:forecastscores}
Our comparison across all three panels uses the frozen donor-trained
tree ensemble. It estimates the boundary fraction separately for each
drain.
Table~\ref{tab:primary-patch} compares the same $5\times5$ patch, basis,
river forcing and drain law under the two coefficient rules. Both pooled
log error and mean absolute discharge error fell on each panel. Across
the three successive panels, 64 of 81 foundations had lower log error
over their twelve candidate drains.

\begin{table}[!ht]
\centering
\caption{Peak seepage discharge at the relief outlet relative to MODFLOW~6 before a
site-specific full-grid check. A foundation improved if the learned
closure lowered log RMSE across its twelve drains.}
\label{tab:primary-patch}
\footnotesize\setlength{\tabcolsep}{2pt}
\begin{tabular}{@{}lrrrrrr@{}}
\hline
 & & \multicolumn{3}{c}{Peak-discharge log RMSE} & \multicolumn{2}{c}{MAE (m$^3$/day)}\\
\cline{3-5}\cline{6-7}
Panel & \shortstack[r]{Foundations\\improved} &
\shortstack[r]{Shared\\closure} & \shortstack[r]{Learned\\closure} & Gain &
\shortstack[r]{Shared\\closure} & \shortstack[r]{Learned\\closure}\\ \hline
Initial & 28/36 & 0.02180 & 0.01610 & \textbf{26.1\%} & 0.168 & 0.113 \\
Confirmation & 20/27 & 0.02904 & 0.02007 & \textbf{30.9\%} & 0.143 & 0.098 \\
Boundary comp. & 16/18 & 0.03138 & 0.01801 & \textbf{42.6\%} & 0.230 & 0.123 \\
\hline
\end{tabular}

\end{table}

Resampling whole foundations gave paired 95\% intervals for the
site-specific minus shared $R_Q$ of $[-0.00975,-0.00162]$ on the initial
panel, $[-0.01334,-0.00377]$ on confirmation and $[-0.01762,-0.00851]$
on the boundary-comparison panel. Figure~\ref{fig:paired} shows the variation
across individual foundations in the first two panels.

\begin{table}[!t]
\centering
\caption{Forecasts before the first site check at 324 candidates from 27 confirmation foundations, each driven by one flood. Head benefit is measured at the toe.
Below, bins use native peak flow (m$^3$/day). Gain is each bin's share
of the net reduction in summed squared log error from learning the patch
coefficient.}
\label{tab:pred}
\small\setlength{\tabcolsep}{4pt}
\begin{tabular}{lrrr}
\hline
 & Peak flow & Flow MAE & Benefit MAE\\
Model & log RMSE & (m$^3$/day) & (mm)\\ \hline
Shared patch &0.02904 &0.143 &0.625\\
Site-conditioned patch &0.02007 &0.098 &0.411\\
Shared ring &0.03651 &0.175 &0.751\\
Neural ring &0.02656 &0.141 &0.622\\
Direct regression &0.02721 &0.107 &0.443\\ \hline
\end{tabular}
\par\smallskip
{\footnotesize\setlength{\tabcolsep}{3pt}
\begin{tabular}{@{}lrrrrr@{}}\hline
Native flow & $[0,1)$ & $[1,5)$ & $[5,10)$ & $[10,20)$ & $[20,\infty)$\\
Sites & 110 & 48 & 49 & 99 & 18\\
Gain (\%) & 0.203 & 60.503 & 13.243 & 22.941 & 3.110\\ \hline
\end{tabular}}
\end{table}
The neural ring improved over the shared ring on 17 of 27 confirmation
foundations. Its pooled error fell by 27.3\% (95\% paired interval for neural
minus shared $R_Q$, $[-0.01574,-0.00285]$). Its interval against direct regression
included zero ($[-0.00408,0.00376]$). The learned patch had the lowest
errors in Table~\ref{tab:pred}, with more fitting foundations and
six local shifted solves per site. Within the matched ring, the
reduction was 18.3\% on the standard foundations and 38.9\% on the
nine with shifted conductivity and storage. Materials and assigned
floods varied together across these groups.

Small flows did not account for the patch result. All 1,512 installed
candidates across the five panels had positive native peak discharge.
On confirmation, the smallest bin contributed less than 1\% of the
log-error reduction, while flows from 1 to 5~m$^3$/day accounted for
about three fifths (Table~\ref{tab:pred}). The choice of metric
explains why the direct regressor and neural ring change rank.
The direct regressor accumulated less absolute error in every bin
starting at 5~m$^3$/day. Its additional squared log error below that
threshold outweighed its gain above it.

\subsection{What the coefficient fit learned on donor foundations}
\label{sec:donor-sensitivity}
Model selection on donors depended on how the boundary coefficient
was scored. At the nearest precomputed fraction the original tree
fit edged ridge by less than 0.0001 in five-horizon loss. At the
continuous fractions used for deployment, ridge had the lower loss
(0.096512 versus 0.097109). The two additional tree seeds (732 and
733) fell on opposite sides of ridge on the grid, and both scored
worse than ridge at continuous fractions. The original tree fit
remained frozen for all three reported patch panels.

Figure~\ref{fig:donor} shows the held-out-donor development comparison.
The optimum shared fraction was 0.0911. On donor sites the grid-optimal
fractions had an interquartile range of 0.076--0.110 and extended up to
1. The estimated tree fractions were more concentrated, ranging from
0.063 to 0.165. Their narrower range is relevant when interpreting
them as approximations to a response target. The matched control tests the value of letting this fraction vary by site.

An archived hindsight comparison measures how much of the six-state
teacher response a single patch coefficient could recover at the 432
initial-panel candidates. At each site we selected the best of the 25
fractions in Eq.~\eqref{eq:label} using that site's teacher response.
For the shared coefficient, predicted coefficient and hindsight fit,
respectively, the root-mean-square log error over five teacher-response
horizons was 0.11854, 0.09518 and 0.07974. Scoring the resulting peak
discharges against MODFLOW gave $R_Q$ of 0.02180, 0.01610 and 0.00722.
The hindsight fit requires the teacher at every test candidate. Its
objective is the response at five horizons, so its discharge error
serves as a diagnostic rather than a lower bound on $R_Q$.

\subsection{Does resolving a wider exterior account for the gain?}
\label{sec:patch-exterior-comparison}

On the prespecified $5\times5$ patch comparison, ridge had lower
log RMSE than the frozen trees on the new foundations
(Table~\ref{tab:patch-radius}). Changing the trees' training target
from cumulative withdrawal response to peak discharge on two donor
floods reduced error on the new panel by 5.6\% and increased it on
confirmation by 2.1\%. The pulse target can be generated independently
of the hydrograph and remained competitive on the tested floods.

\begin{table}[!t]
\centering
\caption{Discharge forecasts before hydraulic checks. The log RMSE
columns pool 324 candidate drains on the archived panel and 216 on
the new panel. MAE is in m$^3$/day on the new panel.}
\label{tab:patch-radius}
\footnotesize
\begin{tabular}{lrrr}
\hline
Patch and boundary rule & Confirmation & Boundary comp. & MAE \\
\hline
$5\times5$ shared       & 0.02904 & 0.03138 & 0.230 \\
$5\times5$ trees        & 0.02007 & 0.01801 & 0.123 \\
$5\times5$ ridge        & 0.01698 & 0.01517 & 0.109 \\
$7\times7$ coarse exterior & 0.05857 & 0.06659 & 0.607 \\
$7\times7$ refitted shared & 0.02506 & 0.02808 & 0.192 \\
$7\times7$ ridge           & 0.01457 & 0.01259 & 0.095 \\
\hline
\end{tabular}
\end{table}

The wider-patch experiment asked whether explicitly resolving an
approximate exterior could account for the improvement. The exploratory
$7\times7$ ridge fit had the lowest error. Against a shared coefficient
recalibrated for the same radius and direct local solver, ridge reduced
log RMSE by 55.1\% on the new panel and 41.8\% on confirmation.
It improved 15 of 18 new foundations (paired difference $-0.01548$,
descriptive 95\% interval $[-0.02145,-0.00852]$). The deterministic
coarse exterior had 5.29 times the ridge error on the new panel.
That comparison changes the exterior treatment. Widening the patch
also changes the local solve from reduced to direct. The matched
comparisons within each radius isolate the effect of site conditioning.
Intervals resample whole foundations 10,000 times at fixed fits without
multiplicity adjustment.

\subsection{Does the ring need a neural coefficient estimator?}
\label{sec:ring-estimators}
Holding the hydraulic ring fixed, the neural estimator had the lowest
pooled peak-discharge log RMSE on the fresh panel (Table~\ref{tab:ring-abl}).
Both the neural and tree rules improved over shared coefficients.
Their respective 95\% paired intervals for the difference from shared
$R_Q$ were $[-0.01502,-0.00265]$ and $[-0.01245,-0.00262]$.

\begin{table}[!t]
\centering
\caption{Frozen ring on 36 new foundations. Discharge errors pool 432 sites.
A placement passes when native discharge meets capacity and peak head is
within 1~mm of the best feasible action (144 decisions per budget).}
\label{tab:ring-abl}
\footnotesize
\begin{tabular}{lrrrr}
\hline
 & Log & Flow MAE & \multicolumn{2}{c}{Placement passes}\\
Estimator & RMSE & (m$^3$/day) & 0 checks & 3 checks\\
\hline
Shared & 0.03390 & 0.250 & 135 & 144\\
Linear (ridge) & 0.02991 & 0.155 & 133 & 144\\
Boosted trees & 0.02622 & 0.180 & 136 & 144\\
Archived neural & 0.02477 & 0.153 & 131 & 144\\
\hline
\end{tabular}
\end{table}

\begin{figure*}[!t]
\centering
\includegraphics[width=.88\textwidth]{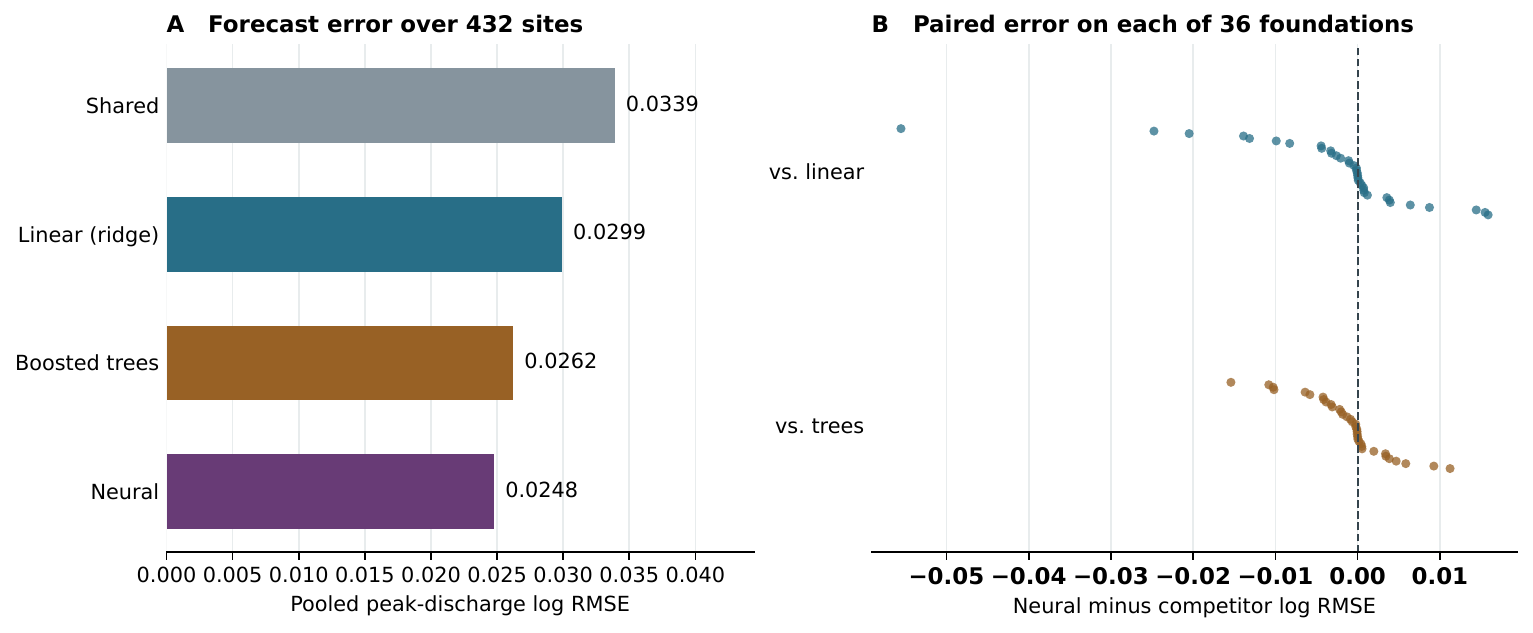}
\caption{Estimating three coefficients on the same hydraulic ring.
(A) Peak-discharge log RMSE across the 432 candidate placements.
(B) Each dot gives the difference in log RMSE across the twelve
candidates on one foundation. Values to the left of zero favor the
original neural head. The bootstrap intervals in the text refer to
differences in pooled RMSE.}
\label{fig:ring-estimator-pairs}
\end{figure*}

Figure~\ref{fig:ring-estimator-pairs} shows the variation across foundations. The neural head had lower error on 20 of 36 foundations compared
with ridge and 23 of 36 compared with trees. The adjusted 97.5\%
intervals for the differences in pooled log RMSE were
$[-0.01313,0.00134]$ versus ridge and $[-0.00433,0.00127]$
versus trees. Both included zero. The prespecified criterion for
a neural advantage over both simpler heads was unmet. The two
additional neural initializations yielded 0.02876 and 0.02610 log
RMSE, on opposite sides of the tree estimate of 0.02622.
Mean absolute peak toe-head benefit errors were 0.621~mm for
the archived network and 0.618~mm for ridge.

At zero checks, the neural ring's 13 failures comprised four
over-capacity selections and nine feasible selections more than 1~mm
above the optimum. In four of those nine the neural forecast excluded
the native optimum as over capacity. In the other five the optimum
remained eligible and the policy preferred another action. Both feasibility
and benefit ranking therefore contributed to the lower pass count.
Boosted trees also selected four over-capacity drains, but made only four
additional errors on feasible actions.

\subsection{How a forecast gain changes a placement}
\label{sec:decisions}
On the initial panel, two checks sufficed for both patch policies to
pass every near-capacity decision. On confirmation, site conditioning
increased the number of acceptable unchecked placements, yet also
increased the number exceeding capacity (Table~\ref{tab:decision}).
After three checks both patch policies met capacity throughout. Both
passed all 81 ordinary-capacity decisions.

To locate errors near the hard constraint, we retrospectively grouped
candidate--capacity pairs by the native margin $|Q_s-C|/C$.
All false admissions by the shared and learned confirmation patches fell
inside the 5\% band. Of the shared patch's 18 false exclusions, 16 were
inside the band and two outside. Every learned-patch false exclusion fell
inside it. The learned patch admitted fewer over-capacity candidates
across the full panel, yet more of them were selected
(Table~\ref{tab:margin}).

\begin{table}[!ht]
\centering\footnotesize
\caption{Unchecked feasibility errors near the capacity limit ($|Q_s-C|\le 0.05C$). Counts refer to candidate--capacity
pairs. The band contained 158 of 1,296 confirmation pairs (78 over capacity)
and 168 of 1,728 matched-ring pairs (82 over capacity). The last column
counts false admissions that became placements.}
\label{tab:margin}
\begin{tabular}{@{}llrrr@{}}\hline
Panel & Closure & False admit & False exclude & Selected\\ \hline
Confirmation & Shared patch & 9 & 16 & 1\\
             & Learned patch & 6 & 12 & 4\\
Matched ring & Shared ring & 6 & 34 & 1\\
             & Neural ring & 6 & 20 & 4\\ \hline
\end{tabular}
\end{table}

On the matched-ring panel, both rules falsely admitted six pairs, all
inside the 5\% band. Three pairs were common to both rules. Every
false admission unique to the neural ring won its head-benefit ranking.
None unique to the shared ring was selected. The identity of the
misclassified site therefore explains why the policies incurred
different violations at the same false-admission count. Each site
forecast was reused at four capacities.

\begin{table}[!t]
\centering
\caption{Near-capacity confirmation decisions, 108 total. A feasible action can still fail the head criterion. The zero-check rule was scored retrospectively. Entries give counts.}
\label{tab:decision}
\small
\begin{tabular}{lrrrr}
\hline
 & \multicolumn{2}{c}{Zero checks} & \multicolumn{2}{c}{Three checks}\\
Model & Pass & Over cap & Pass & Over cap\\ \hline
Shared patch & 97 & 1 & 106 & 0\\
Site-conditioned patch & 103 & 4 & 107 & 0\\
Shared ring & 94 & 2 & 107 & 0\\
Neural ring & 100 & 4 & 107 & 0\\
Direct regression & 99 & 7 & 107 & 0\\ \hline
\end{tabular}
\end{table}

\begin{figure*}[!t]
 \centering\includegraphics[width=.94\textwidth]{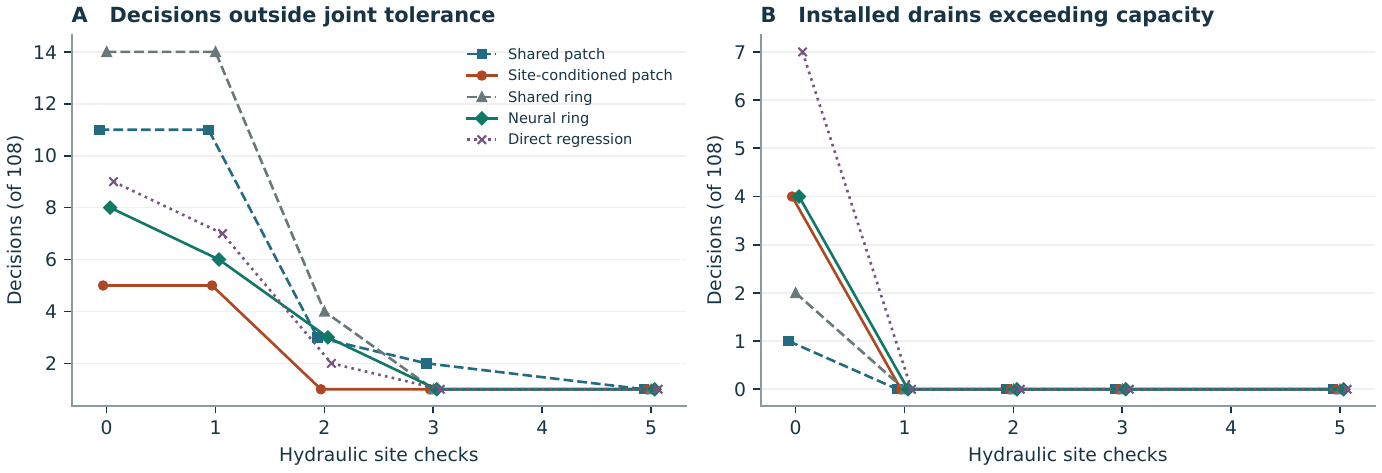}
 \caption{Confirmation decisions under the four near-capacity limits at
 each of 27 foundations. (A) Count failing either native feasibility or the
 1~mm toe-head tolerance. (B) Count exceeding native discharge capacity.
 Zero checks is a retrospective ranking of the saved forecasts. The other
 budgets replay frozen query sequences. Four checks were untested. Markers are offset horizontally
 within each integer budget to reveal coincident counts.}
 \label{fig:budget}
\end{figure*}

Figure~\ref{fig:budget} shows how checks resolved capacity errors on
confirmation. The remaining failure for the site-conditioned patch
after three checks was a feasible placement with too little head relief.

\begin{table}[!t]
\centering\footnotesize\setlength{\tabcolsep}{3pt}
\caption{Placement replays at near-capacity limits. Parentheses give
foundations, $n$ gives decisions. The first two panels compare shared
and ridge coefficients on the same $7\times7$ patch. The last compares
shared and neural coefficients on the same ring. Pass requires native
feasibility and peak toe head within 1~mm of optimum. Over counts
capacity violations. Exact counts recovery of the native choice.
All three-check decisions shown met capacity.}
\label{tab:placement-replay}
\begin{tabular}{@{}llrrrrr@{}}\hline
 & & & \multicolumn{2}{c}{Zero checks} & \multicolumn{2}{c}{Three checks}\\
\cline{4-5}\cline{6-7}
Panel & Rule & $n$ & Pass & Over & Pass & Exact \\ \hline
Boundary (18) & Shared & 72 & 67 & 0 & 72 & 72 \\
 & Ridge & 72 & 69 & 1 & 72 & 72 \\
Confirm. (27) & Shared & 108 & 102 & 2 & 106 & 106 \\
 & Ridge & 108 & 101 & 5 & 107 & 107 \\
Ring (36) & Shared & 144 & 135 & 1 & 144 & 144 \\
 & Neural & 144 & 131 & 4 & 144 & 143 \\
\hline\end{tabular}

\end{table}

The wider-patch replays used the same GP uncertainty parameters and
checker for shared and ridge coefficients. Both recovered every native
choice after three checks on the boundary-comparison panel. On
confirmation, ridge recovered one more (Table~\ref{tab:placement-replay}).
At that budget, every ring estimator met the joint tolerance. The
three site-conditioned ring estimators missed the same exact choice
by 0.264~mm of head relief. The shared ring recovered it.

\subsection{A small flow error can exclude the most effective drain}
\begin{figure*}[t]
 \centering\includegraphics[width=.96\textwidth]{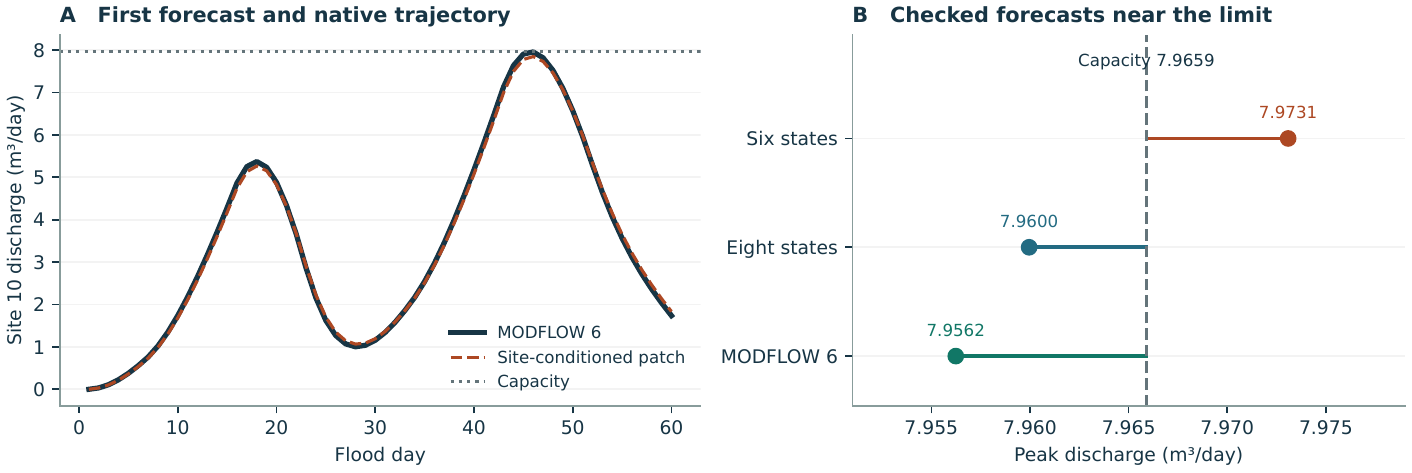}
 \caption{Boundary case on foundation 3128, site 10. (A) Daily native
 candidate discharge and the pre-check forecast from the saved site-conditioned
 patch. (B) Enlarged view of three peak estimates against the identical
 capacity, 7.9659~m$^3$/day. The six-state estimate excluded the site.
 The eight-state estimate placed it on the same side as the native peak.
 The horizontal axes in the two panels measure different quantities.}
 \label{fig:borderline}
\end{figure*}

A small checking error mattered most at a site with substantial head
relief and almost no margin below capacity. At the site highlighted
in Figs.~\ref{fig:decision-loop} and~\ref{fig:borderline}, the
six-state check overestimated the native peak by
0.01685~m$^3$/day. The native peak lay only 0.00966~m$^3$/day below
the limit. Both the learned patch and neural ring therefore chose
site 9, whose native peak toe head was 16.97~mm higher than at site
10. The retrospective eight-state recheck restored site 10 to the
neural ring's feasible set and recovered the native choice.

The archived refinement covers 101 distinct confirmation sites across
27 foundations, selected because a checked forecast was within 5\% of
a capacity. On this subset the RMS difference in stabilized log peak
flow was 0.01920 between the site-conditioned patch and six-state
checker, 0.00071 between the checker and MODFLOW, and 0.00015 between
the eight-state recheck and MODFLOW. Patch-to-MODFLOW error was 0.01924.
The patch and checker share their river and toe responses, so their
difference isolates the effect of replacing the local response within
this rollout. The checker-to-MODFLOW difference includes errors in the
common responses and the site approximation. These RMS distances do
not add. The subset was selected for refinement; its errors describe
the checked sites close to a limit.

\begin{figure*}[!t]
\centering\includegraphics[width=.95\textwidth]{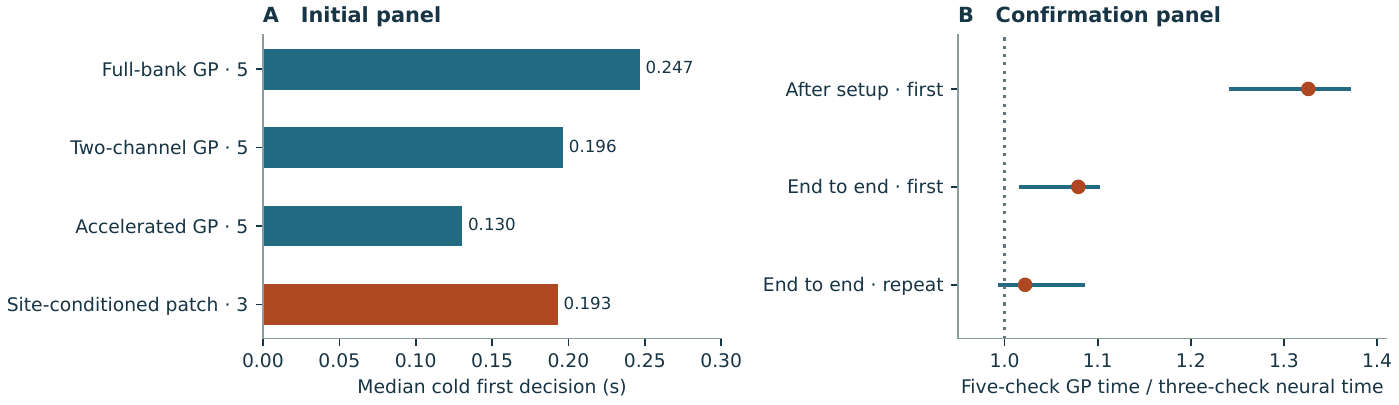}
\caption{Wall time under the matched cold timing comparisons. (A) The
initial panel's median first decision includes setup. The number after
each method is its ordinary check budget. The accelerated GP and
site-conditioned patch share the optimized hydraulic backend. (B) Points
are foundation-paired ratios and bars are descriptive 95\% foundation
intervals. Each confirmation series uses the same nine foundations and four paired repetitions per foundation. Values above one favor the three-check neural policy. The
after-setup ratio excludes the common foundation build.}
\label{fig:timing}
\end{figure*}

\subsection{A frozen recheck on a new boundary challenge}
Every policy met capacity and the 1~mm joint tolerance on the fresh
challenge. Exact choices reveal the limit of refinement
(Table~\ref{tab:challenge}). Rechecking recovered the native optimum
throughout for the shared ring, direct regressor and five-check GP.
The neural ring had left the same candidate off its shortlist under
three capacities on one foundation. Selecting the neighboring candidate
cost 0.757~mm in peak head. A more accurate check could change a
shortlisted candidate's eligibility. It could not recover the omitted
candidate.

\begin{table}[t]
\centering\caption{Fresh challenge, 36 near-limit decisions. Columns show
ordinary checks per decision, exact choices before and after the eight-state
recheck, and total extra site recalculations across decisions.}
\label{tab:challenge}\small
\begin{tabular}{lrrrr}\hline
Prior & Checks & Before & After & Extra\\ \hline
Shared ring & 3 & 34 & 36 & 50\\
Neural ring & 3 & 32 & 33 & 47\\
Direct regression & 3 & 34 & 36 & 51\\
Accelerated GP & 5 & 34 & 36 & 65\\ \hline
\end{tabular}
\end{table}

\subsection{Cost of a placement}
The initial three-check patch matched the five-check accelerated GP
on placement passes but took longer per cold first decision
(Fig.~\ref{fig:timing}A). Both used the same hydraulic backend.
The full-bank and exact two-channel timing controls show the cost of
preparing the common responses before a site is checked.

On confirmation, the three-check neural ring and five-check GP again
had the same pass count. Timing favored the ring more strongly after
setup than over a complete decision. In the repeat series, the
end-to-end GP-to-ring time ratio was 1.022 (descriptive 95\% foundation
interval 0.996--1.084). In the supplemental boundary challenge,
where both recovered every optimum after refinement, the corresponding
GP-to-shared-ring ratio was 1.064 (1.021--1.094), including rechecks.
These timings describe the 6,000-cell grids on one CPU implementation.

Site conditioning also carries a cost when every drain is screened.
On nine standard boundary-comparison foundations, preparing the common
responses and forecasting twelve sites took a median 0.212~s with
$7\times7$ ridge and 0.133~s with its shared-coefficient control,
before any site checks.

\section{Discussion}
\subsection{What the local coefficient captures}
Equation~\eqref{eq:schur} identifies what the local patch loses.
At each hydraulic shift the exterior contributes a matrix that can couple
different perimeter cells. For the scalar closure in Eq.~\eqref{eq:patch}, set
$R=(J_s(\alpha)+\sigma I)^{-1}$ at $\sigma>0$.
Differentiating the reduced withdrawal response gives
\begin{equation}
  \frac{\mathrm d}{\mathrm d\alpha}(v_s^\top Rv_s)
  = -(Rv_s)^\top B_s^{\rm cut}(Rv_s)\leq 0 .
  \label{eq:closure-monotonicity}
\end{equation}
The inequality follows because $B_s^{\rm cut}$ is the projection of
a nonnegative diagonal matrix of severed conductances. Increasing the
fraction therefore decreases or leaves unchanged the local withdrawal
response at that shift. On the fitted interval $\alpha\in[0.012,1]$,
continuity guarantees a match for any scalar target between the two
endpoint responses. This monotonicity is a property of the shifted local
response alone. Peak discharge, placement error and agreement with
MODFLOW~6 under a flood hydrograph require separate evaluation. The donor labels summarize five time horizons.
The resulting coefficient identifies an effective response for this outlet.
Individual exterior conductances remain unidentified.

The operator target separates the site from the event. Under the same
geometry, material properties and time step, a new flood changes the
driving head and hence the discharge history. It requires no new
exchange coefficients. The original coefficient estimators see material,
geometry and local response descriptors. Floods do enter training of
the GP uncertainty model and the direct-output comparator. Performance
on new floods remains an empirical question.

Within each matched patch, estimating exchange by site improved the
forecast. Across the two patch sizes, the local solver also changed.
The ring experiment held the hydraulic representation fixed. Its
adjusted intervals left the relative performance of neural, ridge and
tree estimators unresolved. Taken together these comparisons identify a
transferable prediction target within this foundation family. They do
not determine an optimal discretization or regressor.

\subsection{What controls a placement error}
The initial panel provides a separate check on the two common response
channels. Across the 36 foundations, the largest relative Euclidean
errors of the accelerated response arrays over candidate sites and lags
were $2.74\times10^{-4}$ for river loading and $3.62\times10^{-4}$ for the
reciprocal toe input, relative to direct full-grid propagation. These
are response-array discrepancies. An all-site replacement of the common
responses on confirmation would be needed to attribute their share of
peak-discharge error.

On foundation 3128, the check misjudged a candidate near capacity.
On 3708, the policy omitted a contender from the checked set. A more
accurate check can correct the first error. The second depends on which
candidates receive the limited checks. Conversely, a better forecast at
a drain already within capacity can leave its rank unchanged. The shared
ring recovered all 144 optima at the three-check budget despite its
larger average forecast error.

\subsection{Scope and external validation}
All 126 held-out foundations are synthetic and belong to related design
families. The stress cases alter conductivity and storage together.
Each foundation has one assigned hydrograph, twelve candidate screens,
a known opening and a prescribed drain conductance. The resampling
unit is a foundation with its assigned flood. Effects of site,
flood history and material class cannot all be separated. Near-limit
capacities were deliberately enriched. The experiments estimate numerical
agreement under the specified model, discretization and scoring rule. The 1~mm criterion compares two predictions at that resolution. Its interpretation as physical accuracy would also require spatial and temporal refinement and independent observations.
Chen et al.~\cite{chen} calibrated the Profit Island levee model to
piezometric heads and relief-well discharge during the 1997 flood.
The corresponding field test would predict both outputs from recorded
stage during a withheld flood. Our simulations quantify hydraulic
loading under a prescribed cutoff condition. Erosion and stability
would require additional models.

Moving from a prescribed foundation to a surveyed reach also requires
a model of the river geometry and a way to link monitoring records to
that geometry. Abdelguerfi's edited volume~\cite{abdelguerfi3d} describes
how geospatial data are assembled into three-dimensional environments.
For the river channel, Flanagin et al.~\cite{flanaginSplines} interpolated
surveyed cross sections into a mesh that could be joined to surrounding
terrain. Two precedents address the data handoff. Tu et
al.~\cite{tuGis} used an adapter to give geographic information systems
access to raster data held in another system. Wilson et
al.~\cite{wilsonGidb} described geographic data import and export
through the GIDB system using XML. For a field test of the closure,
the hydraulic observations would still have to be matched to the
modeled locations and flood dates.

Timings include online setup on one grid size. A scaling test would
measure cold and repeated forecasts on larger grids. An adaptive
exterior could resolve conductivity contrasts within the present
one-head-per-layer aggregates.

\section{Conclusion}
SCBC improved forecasts of peak seepage discharge captured at unseen
relief outlets across three panels of synthetic levee foundations, with
MODFLOW~6 as the reference. The learned coefficient represents how the
omitted foundation affects drawdown at the outlet. At fixed hydraulic
properties and time step, that response can be reused across flood
hydrographs.

For screening every candidate, the results favor site-conditioned
exchange. On the nine confirmation foundations with altered conductivity
and storage, the neural ring reduced peak-discharge log RMSE by
38.9\% relative to the shared ring. In a separate exploratory
comparison, the deterministic coarse exterior had 5.29 times the
log RMSE of the $7\times7$ ridge patch. Each comparison applies to
its tested foundation panel and exterior treatment.

For placing one relief outlet to minimize peak toe head under a hard
discharge limit, a shared ring
with three site checks recovered all 144 native optima on the
matched-ring panel. On the separate challenge it recovered 34 of 36
with three ordinary checks and all 36 after selective eight-state
rechecks. Refinement can resolve a borderline candidate only if the
policy has checked that site. For this task, both the shortlist and
the accuracy of its near-capacity checks matter. A field test would compare both peak outlet discharge and peak toe head
with measurements during a flood withheld from calibration.

\bibliographystyle{IEEEtran}
\bibliography{references}
\end{document}